\def\year{2022}\relax
\documentclass[letterpaper]{article} 
\usepackage{aaai22}  
\usepackage{times}  
\usepackage{helvet}  
\usepackage{courier}  
\usepackage[hyphens]{url}  
\usepackage{graphicx} 
\usepackage{natbib}  
\usepackage{caption} 
\DeclareCaptionStyle{ruled}{labelfont=normalfont,labelsep=colon,strut=off} 
\usepackage{algorithm}
\usepackage{algorithmic}

\usepackage{newfloat}
\usepackage{listings}
\floatstyle{ruled}
\newfloat{listing}{tb}{lst}{}
\floatname{listing}{Listing}
\usepackage{microtype}
\usepackage{graphicx}
\usepackage{color,hyperref}
\usepackage{amsmath,amssymb}
\usepackage{amsmath}
\usepackage{lipsum}
\usepackage{subcaption}
\usepackage{comment}
\usepackage{url}
\usepackage{multirow}
\usepackage{adjustbox}                          
\usepackage{adjustbox}
\usepackage{multirow}
\usepackage{booktabs}
\DeclareUnicodeCharacter{2212}{-}

\title{Twitter User Representation Using Weakly Supervised Graph Embedding}
\author{
    Tunazzina Islam,
    Dan Goldwasser 
}
\affiliations{
    Department of Computer Science, Purdue University, West Lafayette, Indiana 47907 \\
    islam32@purdue.edu, dgoldwas@purdue.edu
}

\begin{document}

\maketitle

\begin{abstract}

Social media platforms provide convenient means for users to participate in multiple online activities on various contents and create fast widespread interactions. However, this rapidly growing access has also increased the diverse information, and characterizing user types to understand people’s lifestyle decisions shared in social media is challenging. In this paper, we propose a weakly supervised graph embedding based framework for understanding user types. We evaluate the user embedding learned using weak supervision over well-being related tweets from Twitter, focusing on `Yoga’, `Keto diet’.  Experiments on real-world datasets demonstrate that the proposed framework outperforms the baselines for detecting user types. Finally, we illustrate data analysis on different types of users (e.g., practitioner vs. promotional) from our dataset. While we focus on lifestyle-related tweets (i.e., yoga, keto), our method for constructing user representation readily generalizes to other domains.
\end{abstract}

\section{Introduction}
Social media has been rapidly evolved to make inferences about the real world, with application to health \cite{dredze2012social,de2013predicting,pantic2014online}, well-being \cite{schwartz2013characterizing}, politics \cite{o2010tweets}, marketing\cite{gopinath2014investigating}. Over the last decade, we have witnessed a dramatic change in social media microblogging platforms, specifically Twitter. An increasing number of people access Twitter to express opinions, engage with friends, share ideas \& thoughts, and propagate various contents in their social circles.
\subsection{Motivation} 
\begin{figure*}[htbp]
  \centering  
  \includegraphics[width=  \textwidth]{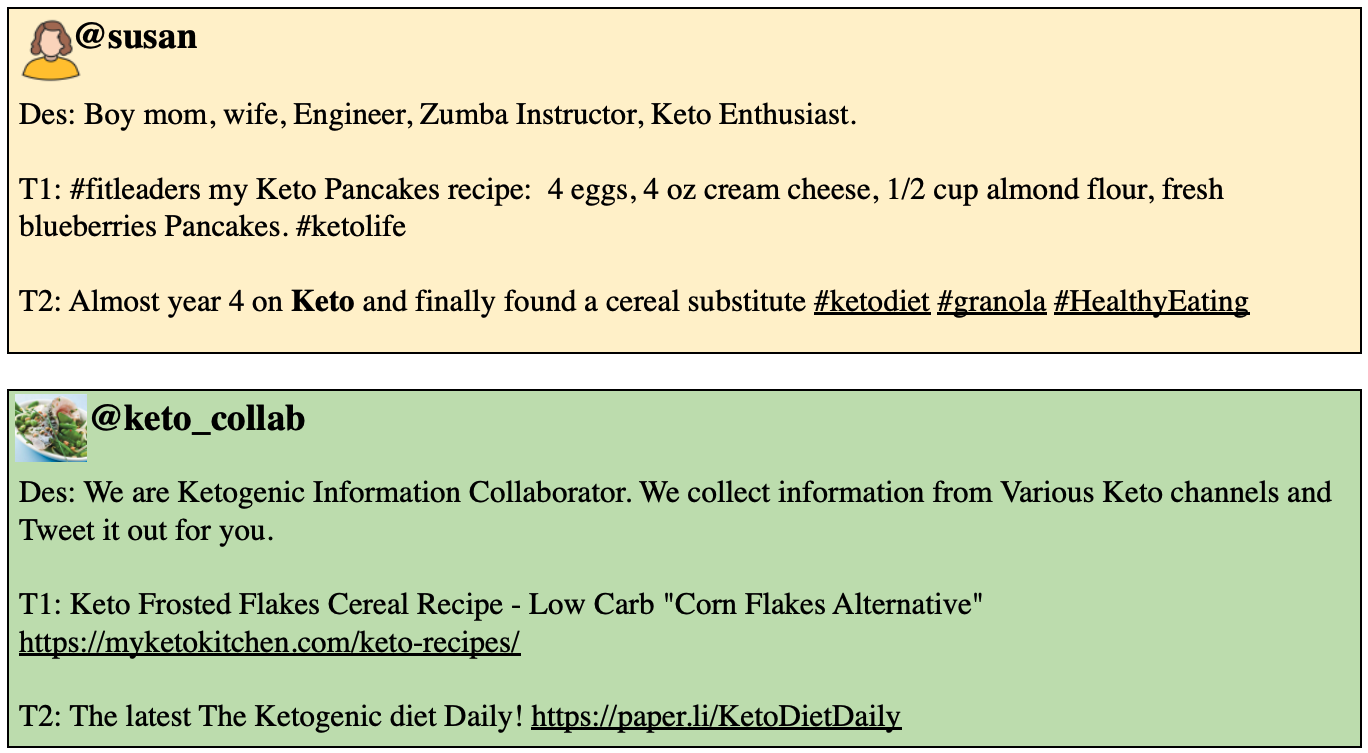}
    \caption{Two different Twitter users' profile sharing their interests on keto diet. Here, \textit{Des} represents users' profile description and $T1, T2$ are users' tweets.}
    \label{fig:keto_user}
\end{figure*}

Twitter is a highly influential and relevant resource for understanding lifestyle choices, health, and well-being \cite{schwartz2016predicting,yang2016life,amir2017quantifying,reece2017forecasting,islam2019yoga}. The information provided is often shaped by people's underlying lifestyle choices, motivations, and interests. Besides, multiple commercial parties, practitioners, and interest groups use this platform to advance their interests and share their journey focusing on specific lifestyle decisions based on different motivations. Fig. \ref{fig:keto_user} shows two different Twitter users' profile description and tweets focusing on keto diet. Though both users \textbf{susan} and \textbf{keto\_collab} tweet about the keto diet, they have different intentions. User \textbf{susan} (yellow box) is a `keto enthusiast' tweeting about her keto journey, while \textbf{keto\_collab} (green box) focuses on collecting information from various keto channels and promoting keto recipes and ketogenic diet daily. In addition to the tweets' contents, the profile description of \textbf{susan} indicates that she is a practitioner. On the other hand, the profile description and tweets of \textbf{keto\_collab} indicate that it is a promotional account (Fig. \ref{fig:keto_user}). Our goal is to automatically classify user types from well-being related tweets and analyze their textual content.

To demonstrate our proposed method to characterize user types, we consider two lifestyle-related activities: \textbf{Yoga} -- a popular multi-faceted activity focusing on the body-mind-spirit connection \cite{goyeche1979yoga} having benefits of alleviating symptoms of anxiety and depression as well as promoting good physical fitness \cite{khalsa2004treatment,yurtkuran2007modified,smith2009evidence,ross2010health} and \textbf{Keto diet} -- a low-carbohydrate, high-fat, adequate-protein diet helping weight loss \cite{johnstone2008effects}, controlling $type-2$ diabetes \cite{mckenzie2017novel} as well as therapeutic potential in pathological conditions, such as PCOS, acne, neurological diseases, cancer, and the amelioration of respiratory and cardiovascular disease risk factors \cite{paoli2013beyond}. Despite the current popularity of yoga and keto, there is little research on analyzing users' lifestyle decision in social media. However, understanding user dynamics on social media is critical to analyze their lifestyle choice and motivation. 

\subsection{Challenges} First, social media users have massive and diverse information regarding topics, content, demographics. It is hard to obtain \textit{large-scale annotated data} for training a machine learning models on such sophisticated content. Second, tweets are \textit{short and often ambiguous}. So, a simple pattern-based analysis on text (tweets) using specific keywords is often inadequate for capturing relevant information. Third, user representation requires \textit{multiview formulation} of data. A system solely relying on tweets' contents misses the rich auxiliary information from the available sources. In addition to the tweet, user's profile description and network are helpful. In this work, we suggest a graph embedding based approach designed to address the above challenges for generating user representation by leveraging weak supervision from profile description.  

\subsection {Existing Work} Prior works infer many latent user characteristics like personality \cite{golbeck2011predicting,kosinski2013private,schwartz2013personality,lynn2020hierarchical}, emotions \cite{wang2015detecting}, happiness \cite{islam2020does}, mental health \cite{amir2017quantifying}, mental disorders \cite{de2013predicting,reece2017forecasting} by analyzing the information in a user's social media account.  
Several works have been done on multiview formulation of data in social media analysis research. \cite{islam2020does} investigated relationship between \textit{practicing yoga} and \textit{being happy} by incorporating textual and temporal information of users using Granger causality. \cite{mishra2018neural,mishra2019abusive} exploited user's community information along with textual features for detecting abusive instances. \cite{ribeiro2018characterizing} characterized hateful users using content as well as user’s activity and connections. \cite{miura2017unifying,ebrahimi2018unified,huang2019hierarchical} used a joint model incorporating different types of available information, including tweet text, user network, and metadata for geolocating Twitter users. These works are typically studied as a supervised learning task.
Previous works aiming to understand Twitter users' types by combining their tweets, metadata, and social information \cite{del2019you,islam2021you,islam2021analysis} rely on large amounts of labeled instances to train supervised models. Such sizeable labeled training data is difficult to obtain. Semi-supervised methods can reduce dependence on labeled texts. Graph based semi-supervised algorithms achieved considerable attention over the years \cite{belkin2006manifold,subramanya2008soft,talukdar2008weakly}. \cite{sindhwani2008document} used a polarity lexicon combined with label propagation \cite{Zhu02learningfrom} for sentiment analysis. Several have used label propagation for polarity classification \cite{blair2008building, brody2010unsupervised}, community detection \cite{jokar2019community}. Our challenge is to construct a user representation in a weakly supervised way relevant for characterizing nuanced activity and lifestyle-specific properties. 

\begin{figure*}[htbp]
  \centering  
  \includegraphics[width=  0.8 \textwidth]{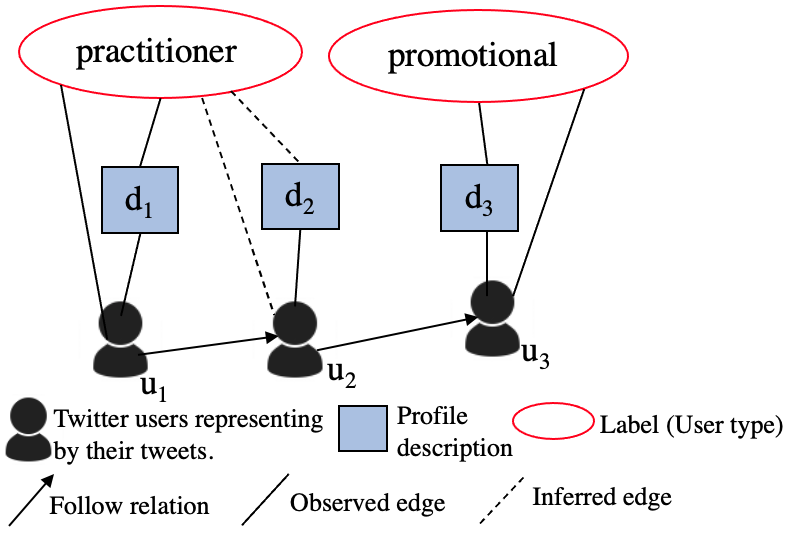}
    \caption{Information graph capturing relations among users, descriptions, user types.}
    \label{fig:user_net}
\end{figure*}

\subsection{Contributions} 
In this paper, our goal is to explore the approach driven by the principle of social homophily \cite{mcpherson2001birds}, indicating individuals' tendency to form social ties with others who have common interests. \cite{yang2017overcoming} used this phenomenon to overcome language variation in sentiment analysis. Graph-based method such as label propagation \cite{Zhu02learningfrom,talukdar2009new,speriosu2011twitter} provides a representation by exploiting such relationships to improve classification, often while requiring less supervision than with standard classification. In our settings, we follow the observation that the users' lifestyle choices expressed in the text will be reflected in users' behavior engaging with it. The main insight of our work is that given two users with very similar behaviors and similar activity, their profile descriptions can be aligned. We define an objective that does the weak supervision - mapping profile descriptions to labels (i.e., user types) based on specific keywords.

Fig. \ref{fig:user_net} represents the relationship between users, descriptions, user types. To capture more hidden relationships
that characterize user types through information graph,
we suggest \textit{inference function} to apply iteratively. Fig. \ref{fig:user_net} shows that users $u_1$ and $u_3$ have weak labels that we obtain from their profile descriptions $d_1$ and $d_3$ respectively by mapping with labels using specific keywords. Profile description of $u_2$ doesn't have label initially. 
Inference function infers the label of user $u_2$ and $u_2$'s description, $d_2$ as `practitioner' because $u_1$ and $u_2$ have similar tweets (Fig. \ref{fig:user_net}). 


From a technical viewpoint, we define our user type detection as a reasoning problem over an information graph that creates distributed representations of nodes contextualized by the graph structure, allowing us to transfer information from observed nodes to unknown nodes (\textbf{Step 1}). \textbf{Step 2} -- we use inference function built on similarity metric defined by the learned graph embedding to increase the number of edges connecting the two node types. We suggest an Expectation–maximization (EM) approach for completing these two steps. Our contributions can be summarized as follows:

\begin{enumerate}
    \item We formulate a novel problem of exploiting weak supervision for user type detection from social media.
    \item We suggest a graph embedding based EM-style approach for learning and reasoning to construct like-minded users incrementally.
    \item  We describe how to generate weak labels from user's profile description along with quantitative quality assessment.
    \item We conduct extensive experiments on real-world datasets to demonstrate the effectiveness of the proposed weakly supervised graph embedding method to detect user type over the baselines.
\end{enumerate}


\section{Model}
In this section, we first describe how to create the information graph and learn the information. Next, we discuss about inference function. Then, we discuss an iterative EM-style approach that continually improves the user representation learned by the graph.

\subsection{Information Graph Creation}
We represent users' activity on social media as an information graph, connecting profile description to user representing by their tweets. In this process, we have following nodes: users representing by their tweets, profile descriptions, user types (practitioner, promotional). We have following observed edges: profile description-to-user type, user-to-user type, profile description-to-user, user-user.
As most users' descriptions are not associated with the labels, our technical challenge is to infer the unknown label of the descriptions. To address this challenge, we learn a graph embedding to maximize the similarity between neighboring nodes \cite{perozzi2014deepwalk,tang2015line,grover2016node2vec}.

\subsection{Information Graph Embedding}
We define the notion of information graph shown in Fig. \ref{fig:user_net}. Let Graph, $G = (V, E)$, where $V$ consists of the following nodes: (1) $D = \{d_1, d_2,...,d_n\}$ represents user's profile description, (2) $U = \{u_1, u_2,...,u_n\}$ are the Twitter users representing by their tweets, (3) $UT=\{ut_1,ut_2,..,ut_m\}$ represents user type (label) which is  binary (i.e., practitioner and promotional) in our case. 
$E$ contains following edges: (1) each Twitter user must have profile description so $D$ connects with $U$, (2) profile description has connection with user type based on specific keywords resulting a connection between $D$ and $UT$, (3) user $U$ connects with user type $UT$ if the user's description is already connected to user type, (4) $U$ and $U$ via follow link. For follow link, we consider those users from our dataset if they are retweeted and/or $@-$mentioned \cite{rahimi2015exploiting} in other users' (from our data) tweets. An edge is created between two users if either user mentioned the other from the data. 
We embed the following instances in a common embedding space - (a) users, (b) profile descriptions, (c) user types. We maximize the similarity between two instances in the embedding space if --
(1) profile description has a type.~
(2) a user has a type.~

Let define the embedding function, $\phi$ that maps the graph's nodes to vectors $R^d$. We train this by following a negative sampling approach. As we construct positive examples by associating each node $u$ with
all of its neighbors, $(u, x^p)$ is a positive pair where $x^{p}$ is a positive example. For each positive pair, we
sample $5$ negative examples ($x^n$) in such a way that 
$u$ and $x^n$ do not share an edge. Our goal is to maximize the similarity of a node embedding with a positive example and minimize the similarity with a negative example. We call a user type a positive example for a user if the user belongs to that type. Otherwise, the type is called a negative example. The embedding loss is designed to place $u$ closer to $x^{p}$ than $x^{n}$. We define the loss function as follows:
\begin{align}
    E_{t}=l(\operatorname{sim}(\phi(u), \phi(x^{p})), \operatorname{sim}(\phi(u), \phi(x^{n})))
\end{align}
where, $E_{t}$ defines the embedding loss for specific objective function $t$ (i.e., user to user type). Similarity function, $sim()$ is the dot product and $l()$ is the cross-entropy loss.
\begin{align}
    l(p,n)=-\operatorname{log}(\frac{e^{\operatorname{sim}(\phi(u), \phi(x^{p}))}}{e^{\operatorname{sim}(\phi(u),\phi(x^{p}))}+e^{\operatorname{sim}(\phi(u),\phi(x^{n}))}})
\end{align}
We minimize the summed loss $\sum_{t \in T} \lambda_{t} E_{t}$, where $T$ is the set of all objective functions and the weight, $\lambda_{t}$ associates with each objective function.

We obtain user embeddings by running a Bi-LSTM \cite{schuster1997bidirectional} over the Glove \cite{pennington2014glove} word embeddings ($300d$) of the words of the user's tweets. Concatenating the hidden states of the two opposite directional LSTMs, we get representation over one time-stamp and average the representations of all time-stamps to obtain a single representation of the tweets. We train this Bi-LSTM jointly with embedding learning. 

\subsection{Inference Function}
After learning the information from graph embedding, which captures the observed relations, we can obtain unobserved relations among nodes in the graph. Based on the similarity captured between non-connected nodes by graph embedding, we create new edges between them to use this knowledge in the future. We do this by defining an inference function that creates the edges based on information graph inferences. For example, we have users whose profile descriptions are not covered by specific keywords used to create a weak label. But those users may have similar tweet contents that match with the tweets of a labeled user. The graph embedding step learn this commonality between these users and represent them with similar embeddings. Using inference function, we directly connect the user and label with an edge as well as the user's description and label with an edge based on the user node similarity in the embedding space.

For the inference function, we make edge connections based
on the node representations learned by computing similarity scores between all pairs of nodes (using the node embedding) and connecting the nodes with the top $k$ scores.

\subsection{EM-style Learning Approach}
In this section, we describe overall EM-style graph
learning framework that continually as follows:

\noindent\textbf{Step 1: Learn node embedding.} In this step, we learn the information graph embedding using the framework described in subsection named Information Graph Embedding to obtain initial graph representation.

\noindent\textbf{Step 2: Infer unlabeled users.} We apply inference function (see details in subsection named Inference Function) built on similarity metric based on the learned information graph representation.

\noindent\textbf{Step 3: Stopping criterion.} At each iteration, after Step 2, we check the predicted labels of all users. Our model converges when the change in predicted labels is less than a threshold between two consecutive iterations.

\section{Dataset Details}
To evaluate our model’s ability to characterize user type, we use two lifestyle choice related data, i.e., yoga and keto diet from Twitter. We download tweets using Tweepy\footnote[1]{\url{https://www.tweepy.org/}} by Twitter streaming API sub-sequentially from May to November, 2019. For yoga, we collect $419608$ tweets related to yoga containing popular keywords : `yoga', `yogi', `yogalife', `yogalove', `yogainspiration', `yogachallenge', `yogaeverywhere', `yogaeveryday', `yogadaily', `yogaeverydamnday', `yogapractice', `yogapose', `yogalover', `yogajourney'. There are $297350$ different users among them $15168$ users have at least yoga-related tweets in their timelines. We have $35392177$ timeline tweets in total. We discard those users who do not have profile description. So we have finally $13301$ yoga users.

For ketogenic diet, we focus on several keywords i.e. `keto', `ketodiet', `ketogenic', `ketosis', `ketogenicdiet', `ketolife', `ketolifestyle', `ketogenicfood', `ketogenicfoodporn', `ketone', `ketogeniclifestyle', `ketogeniccommunity', `ketocommunity', `ketojourney' to extract $75048$ tweets from $38597$ different users. Among them $16446$ users have at least two keto-related tweets in their timelines having total $39250716$ timeline tweets. After discarding the users not having profile description, we have $14320$ keto users finally.

\subsection{Holdout Data} For testing purpose, we manually annotate $786$ yoga users and $908$ keto users using binary label `practitioner', `promotional'. For each user, we check both their profile description and timeline tweets (only yoga/keto-related). At first, we look at the user profile description for user type, whether they explicitly mention practicing a specific lifestyle (e.g., yogi, yogini, keto enthusiast, ketosis); then, we look into the timeline tweets of that user. If the user tweets about the first-hand experience of practicing yoga (e.g., \textit{love doing yoga during weekend morning})/keto diet (e.g., \textit{lost $7lb$ in $3^{rd}$ week of keto dieting}), we annotate them as a `practitioner'.  After looking at the description and tweets, if we observe that they are promoting a gym/studio (e.g., \textit{offering online yoga classes}), online shop (e.g., \textit{selling yoga pants}), app (e.g., \textit{keto diet recipe}), restaurant (e.g., \textit{serving keto meal}) etc., rather than sharing their first-hand experience about a particular lifestyle, we annotate the user as a `promotional' user. 

Two graduate students from Computer Science department manually annotate a subset of tweets ($10\%$) for calculating inter-annotator agreement. This subset has an inter-annotator agreement of $64.7\%$, which is substantial agreement using Cohen’s Kappa coefficient \cite{cohen1960coefficient}. In case of a disagreement, we resolve it by discussion.

\subsection{Information Graph} 
In our information graph, we have following nodes: users representing by their tweets, profile descriptions, user types and following observed edges: profile description-to-user type, user-to-user type, profile description-to-user, user-user.
For yoga data, initially we have $33784$ nodes with $8730$ observed edges. After EM-style approach, we have total $19927$ edges in our information graph for yoga. In our keto data, we have $28583$ nodes and $132486$ observed edges initially. After EM-style approach, we notice that our keto information graph contains total $145944$ edges.


\section{Constructing Weak Labels}
In this section, we describe how to generate weak labels from users’ profile descriptions that can be incorporated as weak sources in our model. We further evaluate the quality of the weak labels.

\subsection{Generating Weak Labels}
\subsubsection{Keyword based knowledge extraction from profile description.}
In some cases, the user's profile description contains specific keywords indicating a particular user type (i.e., practitioner, promotional). We use those keywords to extract the label and utilize them as weak-supervision for our model. However, not to make our weak label too noisy, we do not assume that people with `yoga' or `keto' mentioned in their profile description would be `practitioner' because other keywords that appear with them would be the indicator of being `promotional'. Also, the exact keywords might appear in both user types. So we need to check other keywords appearing with them as separating criteria for `practitioner' and `promotional'. To handle these cases, we form the following rules based on more related keywords:

For yoga, if user description contains following words \textit{(`yoga'  or `yogi'  or `yogini'  or `guru'  or `sprituality'  or `asana'  or `meditation'  or `mindfulness'  or `fitness'  or `health'  ) and not (`studio', `gym', `magazine', `daily', `channel', `clothing', `subscribe', `training', `shop', `sale', `discount', `package', `free', `ship', `retreat', `resort', `design', `jewellery', `handmade', `business', `spa', `hotel', `restaurant', `activewear', `beachwear', `sportsfashion', `festival', `app', `website', `community', `organizer', `center', `donate', `support', `fund', `product', `review', `trend', `healthcare')}, we consider them yoga practitioner. If profile description has following words \textit{(`gym'  or `studio'  or `magazine'  or `daily'  or `channel'  or `clothing'  or `subscribe' or `training'  or `shop'  or `sale'  or `discount'  or `package'  or  `free'  or `ship'  or `retreat'  or `resort'  or `design'  or `jewellery'  or `handmade'  or `business'  or `spa'  or `hotel'  or `restaurant'  or `activewear'  or `beachwear'  or `sportsfashion'  or `festival'  or `app'  or `website'  or `community'  or `organizer'  or `center' or `donate'  or `support'  or `fund'  or `product'  or `review'  or `trend'  or `healthcare')}, we consider them promotional user. After applying keyword based knowledge extraction rule, we have $2104$ weak labels for yoga data.

For keto diet, if user description contains following words \textit{(`keto' or `ketogenic' or `ketosis' or `ketodiet' or `ketogenicdiet' or `lowcarb' or `carnivore', `pro-meat') and (`lifestyle', `follower', `teacher', `instructor', `fitness', `bodybuilder', `nutrition', `coach', `addict', `enthusiast', `health', `body', `mind', `practitioner', `guru', `journey', `life')}, we consider them keto practitioner. On the other hand, if user description has following words
\textit{(`keto' or `ketogenic' or `ketosis' or `ketodiet' or `ketogenicdiet' or `lowcarb') and (`collaborator', `channel', `restaurant', `business', `subscribe', `sale', `discount', `offer', `free', `organizer', `mealplan', `recipe', `trend', `review', `delivery', `app', `community', `website')}, we label them as promotional keto users. Finally, $858$ keto users have weak label.

\subsection{Quality of Weak Labeling}
To assess the weak label quality, we consider the data with both weak label and ground-truth label. For yoga data, we have $451$ users and for keto, we have $56$ users having both weak and true label. We compare the weak labels with the ground truth labels. The accuracy and macro-avg F1 score of the weak label are $0.79$ and $0.78$ respectively for yoga data. For keto data, the accuracy and macro-avg F1 score of the weak label are $0.86$ and $0.67$ correspondingly. We observe that the accuracy and macro-avg F1 score of the weak label both for yoga and keto data are significantly better than random $(0.5)$ for binary classification, indicating that our weak labeling approach has acceptable quality.

\begin{table*}
\centering
\begin{tabular}{lcccc}
    \toprule
    \multirow{2}{*}{\textbf{Model}} & \multicolumn{2}{c}{\textbf{Yoga}} & \multicolumn{2}{c}{\textbf{Keto}} \\
    \cmidrule(r){2-3}\cmidrule(l){4-5}
     & \textbf{Accuracy}  & \textbf{Macro-avg F1}  & \textbf{Accuracy}  & \textbf{Macro-avg F1}  \\
     \midrule
    LSTM\_Glove  &  0.51  & 0.45 &  0.72 &  0.43 \\
    Fine-tuned BERT & 0.47  & 0.47 & 0.72  & 0.42 \\
    Label propagation  &  0.78  & 0.75 & 0.66 & 0.42 \\
    \textbf{EM-style approach} &  \textbf{0.78} & \textbf{0.76} & \textbf{0.72} & \textbf{0.64} \\
    \bottomrule
    \end{tabular}%
\caption{The first two rows are supervised baselines. The last two rows show the user type detection results for weak supervision settings.}
\label{tab:result}
\end{table*}

\begin{table*}
\centering
\begin{tabular}{lcccc}
    \toprule
    \multirow{2}{*}{\textbf{Model}} & \multicolumn{2}{c}{\textbf{Yoga}} & \multicolumn{2}{c}{\textbf{Keto}} \\
    \cmidrule(r){2-3}\cmidrule(l){4-5}
     & \textbf{Accuracy}  & \textbf{Macro-avg F1}  & \textbf{Accuracy}  & \textbf{Macro-avg F1}  \\
     \midrule
    Label propagation (des)  &  0.721  & 0.711 &  0.715 &  0.398\\
    EM-style approach (des) & 0.781  & 0.761 & 0.664  & 0.635 \\
    Label propagation (net)  &  0.573  & 0.572 & 0.644 & 0.384 \\
    EM-style approach (net) &  0.670 & 0.657 & 0.707 & 0.617 \\
    Label propagation (des + net)  &  0.781  & 0.753 & 0.663 & 0.418 \\
    \textbf{EM-style approach (des + net)} &  \textbf{0.782} & \textbf{0.763} & \textbf{0.722} & \textbf{0.642} \\
    \bottomrule
    \bottomrule
    des : profile description \\
    net : user network \\
    des + net : both profile description and user network \\
    \bottomrule
    \end{tabular}%
\caption{Ablation study.}
\label{tab:ablation}
\end{table*}
\section{Experiments}
In this section, we present the experiments to evaluate the effectiveness of our model, baselines, hyperparameter tuning and text pre-processing details. We aim to answer the following evaluation questions (EQ):
\begin{itemize}
    \item \textbf{EQ1:} Can our model improve user type detection task by leveraging weak supervision?
    \item \textbf{EQ2:} How effective to consider multiview information for improving prediction performance?
   \item \textbf{EQ3:} How meaningful our model's embeddings are? How does EM-style approach help?
    
\end{itemize}

\subsection{Experimental Settings}
We use accuracy and macro-average F1 score as the evaluation metrics. We show two versions of our model (weakly supervised) \textemdash (1) initial graph embedding without iterating multiple times (Label propagation), (2) EM-style approach. Both of them are trained on weakly labeled data and evaluated with the performance on holdout data (manually annotated ground-truth discussed in section named Dataset Details). Both of our weakly supervised models are repeated for $3$ times and we report the average performance ($3^{rd}$ and $4^{th}$ rows of Table \ref{tab:result}). We compare our model with two supervised learning baselines. Table \ref{tab:result} shows that our EM-style graph embedding model achieves the highest performance both in accuracy (yoga = $78.2\%$, keto = $72.2\%$) and macro-avg F1 score (yoga = $76.3\%$ , keto = $64.2\%$) than the supervised baselines which answer the \textbf{EQ1}.
\subsubsection{Baselines.} 
For supervised baselines, we train a model using the weak labels assigned by the keywords and make predictions about the users that do not match any keyword. From the weakly labeled training data, we randomly choose $20\%$ data as validation set. Our first supervised baseline is referred to as LSTM\_Glove, where we use $300d$ Glove word embeddings to obtain the embedding of user's tweets and forward these embeddings to LSTM \cite{hochreiter1997long}. We use cross-entropy loss. We fine-tune the pre-trained BERT (base-uncased) \cite{devlin2019bert} model with user's tweets as our second supervised baseline. For both baselines, we use $3$-fold cross-validation and the average is reported in Table \ref{tab:result}.


\subsubsection{Hyperparameter Details.}
We run our proposed model for characterizing user type on yoga and keto data. For both of them, we run the embedding learning (described in subsection named Information Graph Embedding) at most $100$ epochs or stop the learning if the embedding learning loss does not decrease for $10$ consecutive epochs. For inference function, we pick the top $20$ most confident predictions based on majority voting and treat them as labeled users. 
For stopping criterion, we set the threshold to $10\%$. We end up running inference function for $2$ iterations for yoga and $3$ iterations for keto. We use optimizer= Adam \cite{kingma2014adam}, learning rate= $0.001$, batch size = $16$. The single layer Bi-LSTM takes $300d$ Glove word embeddings as inputs and maps to a $150d$ hidden layer. We initialize the embeddings of all of the other instances randomly in $300d$ space. We initialize the weight associates with each objective function, $\lambda_{t}=1$, for all.

For LSTM\_Glove baseline, the single layer LSTM takes $300d$ Glove word embeddings as inputs and maps to a $150d$ hidden layer with optimizer= Adam, learning rate= $0.01$, batch size = $16$, epochs = $20$. For BERT fine-tuning, to encode our texts, for padding or truncating we decide maximum sentence length = $500$. We use batch size = $32$, learning rate = $2e-5$, optimizer= AdamW \cite{loshchilov2018decoupled}, epochs = $3$, epsilon parameter = $1e-8$. Our stopping criterion for these two supervised baselines is the lowest validation loss.

\subsubsection{Pre-processing.}
To pre-process the tweets, we first convert them into lower case, remove URLs, smiley, emoji, stopwords. To prepare the data for input to BERT, we tokenize the text using BERT's wordpiece tokenizer.


\subsection{Does Multiview Information Help? }

Social media user representation requires multiview formulation of data. To understand the contribution of the model components, we perform an ablation study. Table \ref{tab:ablation} shows the results of our ablation study both for yoga and keto data.

We train both initial graph embedding (Label propagation) and EM-style approach with user description only ($1^{st}$ and $2^{nd}$ rows of Table \ref{tab:ablation}). In this case, our information graph contains the following nodes: users representing by their tweets, profile descriptions, user types (practitioner, promotional). We have the following observed edges: profile description-to-user type, user-to-user type, profile description-to-user. For yoga data, initially, we have $26482$ nodes with $2104$ observed edges. After the EM-style approach, we have total $13301$ edges in the information graph of yoga. For keto data, we have $28583$ nodes and $855$ observed edges initially. After the EM-style approach, we notice that our information graph contains total $14313$ edges for keto. Table \ref{tab:ablation} ($2^{nd}$ row) shows that EM-style graph embedding model with description only achieves $78.1\%$ accuracy for yoga and $66.4\%$ for keto as well as $76.1\%$ macro-avg F1 score for yoga and $63.5\%$ for keto.

Then, we train both Label propagation and EM-style approach with user network only ($3^{rd}$ and $4^{th}$ rows of Table \ref{tab:ablation}). Information graph contains the following nodes: users representing by their tweets, profile descriptions, user types (practitioner, promotional) and we have the following observed edges: profile description-to-user type, user-to-user type, user-user. In this case, we have $13304$ nodes and $5811$ observed edges initially for yoga data. After the EM-style approach, we obtain total $6469$ edges in the yoga information graph. For keto data, initially, we have $14316$ nodes with $126929$ observed edges. After the EM-style approach, we have total $131631$ edges in the information graph of keto. The $4^{th}$ row of Table \ref{tab:ablation} shows that the EM-style graph embedding model with user network only obtains $67.0\%$ accuracy for yoga and $70.7\%$ for keto as well as $65.7\%$ macro-avg F1 score for yoga and $61.7\%$ for keto.

Exploiting the knowledge from multiple views to represent the data, user's profile description, and network are helpful besides tweets. In our model, the following instances - tweets, user network, profile descriptions, user types are embedded in a common embedding space which creates distributed representations of nodes contextualized by the graph structure. This graph structure allows us to transfer information from observed nodes to unobserved nodes. The inference function built on similarity metric increases the number of edges connecting the two node types. $5^{th}$ and $6^{th}$ rows of Table \ref{tab:ablation} show two versions of our weakly supervised model i.e., Label propagation and EM-style approach respectively. Our EM-style approach obtains the highest accuracy and macro-avg F1 score both for yoga and keto data ($6^{th}$ row of Table \ref{tab:ablation}). From Table \ref{tab:ablation}, we notice that considering multiview information improves prediction performance (answer to \textbf{EQ2}) compared to the models that are using only either profile description or user network information.

\begin{figure*}
\begin{subfigure}{\columnwidth}
  \centering
  \includegraphics[width=\textwidth]{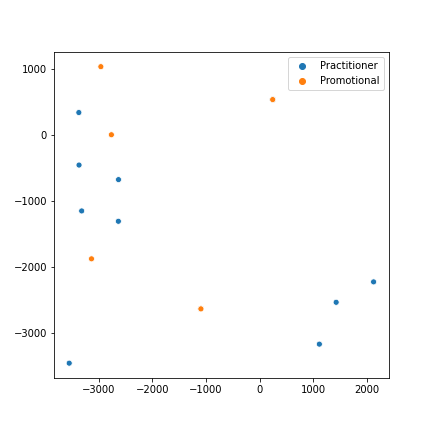}
  \caption{yoga: label propagation}
  \label{fig:graph_learn_yoga}
\end{subfigure}%
\begin{subfigure}{\columnwidth}
  \centering
  \includegraphics[width=\textwidth]{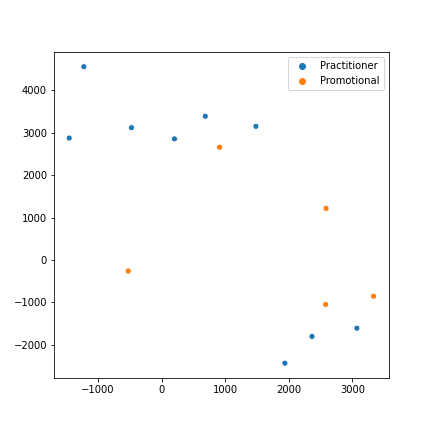}
  \caption{yoga: EM-style}
  \label{fig:em2_yoga}
\end{subfigure}
\begin{subfigure}{\columnwidth}
  \centering
  \includegraphics[width=\textwidth]{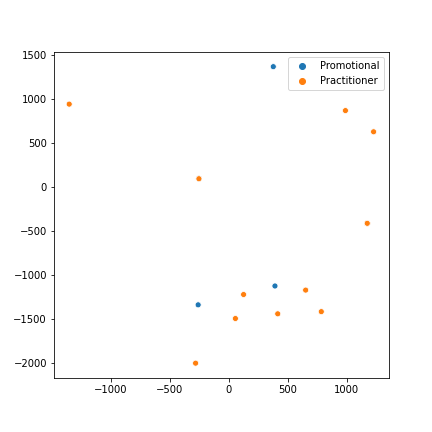}
  \caption{keto: label propagation}
  \label{fig:graph_learn_keto}
\end{subfigure}%
\begin{subfigure}{\columnwidth}
  \centering
  \includegraphics[width=\textwidth]{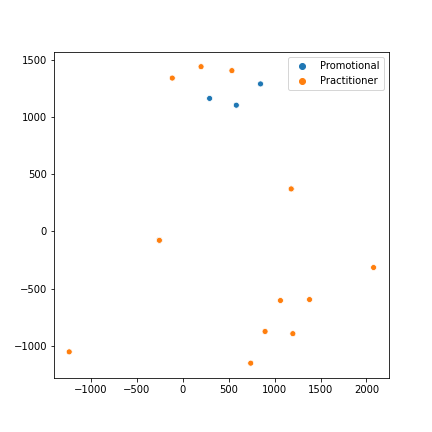}
  \caption{keto: EM-style}
  \label{fig:em3_keto}
\end{subfigure}
\caption{User embeddings after label propagation and EM-style approach both for yoga and keto data in t-SNE visualization.}
\label{fig:embd}
\end{figure*}

\begin{figure}
\begin{subfigure}{1\columnwidth}
  \centering
  \includegraphics[width=\textwidth]{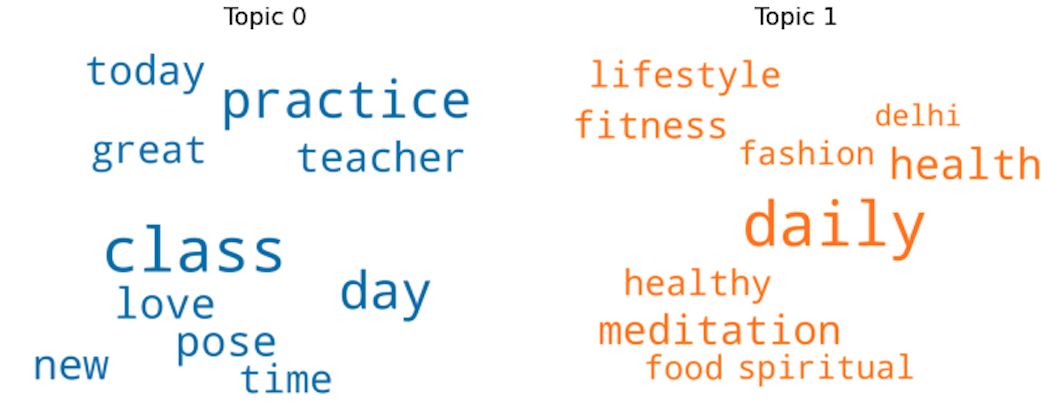}
  \caption{yoga: practitioner}
  \label{fig:prac_yoga_twt_topic}
\end{subfigure}
\begin{subfigure}{1\columnwidth}
  \centering
  \includegraphics[width=\textwidth]{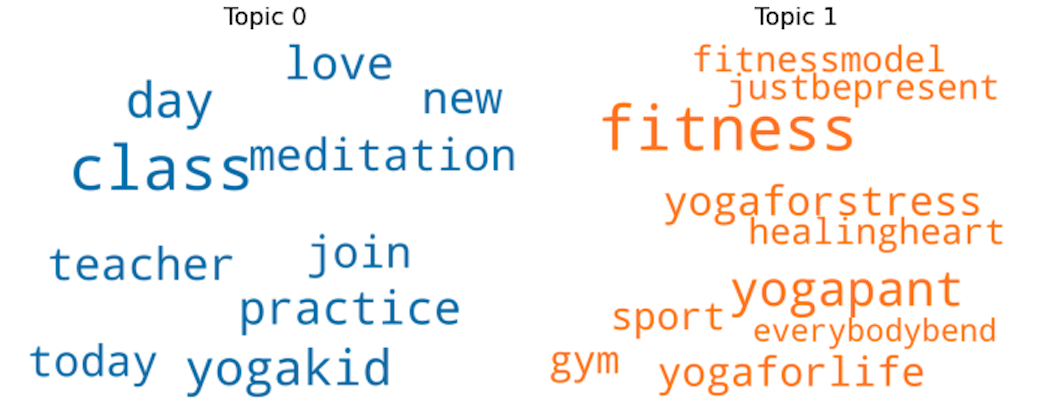}
  \caption{yoga: promotional}
  \label{fig:promo_yoga_twt_topic}
\end{subfigure}
\begin{subfigure}{1\columnwidth}
  \centering
  \includegraphics[width=\textwidth]{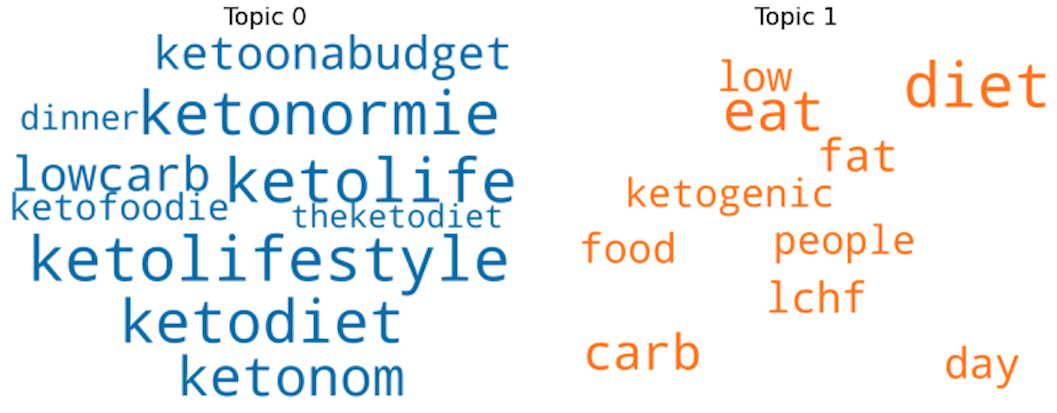}
  \caption{keto: practitioner}
  \label{fig:prac_keto_twt_topic}
\end{subfigure}
\begin{subfigure}{1\columnwidth}
  \centering
  \includegraphics[width=\textwidth]{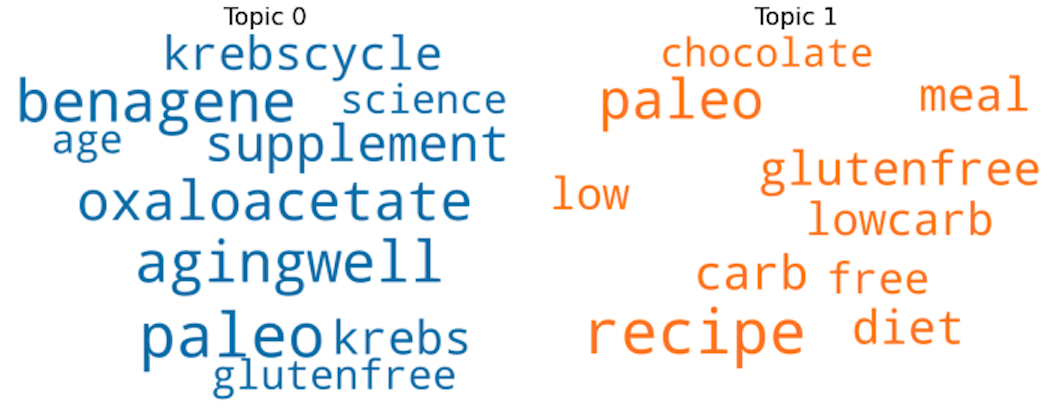}
  \caption{keto: promotional}
  \label{fig:promo_keto_twt_topic}
\end{subfigure}
\caption{Wordclouds of top $10$ keywords in each topic from yoga and keto users' tweets. Only $2$ topics from each dataset are shown here.}
\label{fig:yoga_keto_twt_topic}
\end{figure}

\subsection{Embedding Analysis}
Our model's embeddings are effective to characterize user types. Before our model is trained, user embeddings are random. To answer the \textbf{EQ3}, we project the embeddings to a $2D$ space using t-SNE \cite{van2008visualizing} and use different colors to represent labels (Fig. \ref{fig:embd}). We plot the user embeddings after label propagation (Fig. \ref{fig:graph_learn_yoga} and Fig. \ref{fig:graph_learn_keto}) and EM-style approach (Fig. \ref{fig:em2_yoga} and Fig. \ref{fig:em3_keto}) based on ground truth data. The embeddings show that EM-style approach has better user representation both for yoga (Fig. \ref{fig:em2_yoga}) and keto (Fig. \ref{fig:em3_keto}) dataset.

\begin{figure}
\begin{subfigure}{.5\columnwidth}
  \centering
  \includegraphics[width=\textwidth]{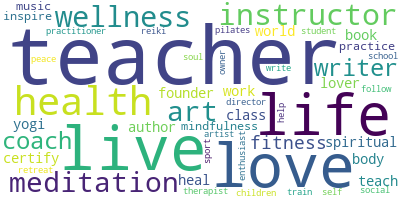}
  \caption{yoga: practitioner}
  \label{fig:prac_yoga_des}
\end{subfigure}%
\begin{subfigure}{.5\columnwidth}
  \centering
  \includegraphics[width=\textwidth]{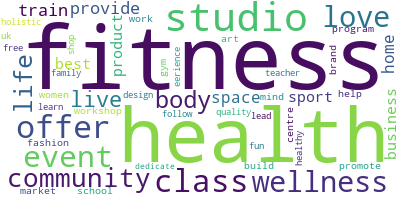}
  \caption{yoga: promotional}
  \label{fig:promo_yoga_des}
\end{subfigure}
\begin{subfigure}{.5\columnwidth}
  \centering
  \includegraphics[width=\textwidth]{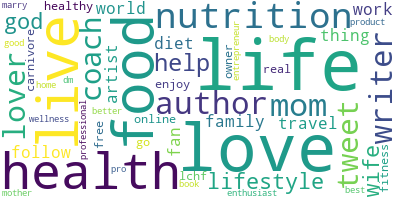}
  \caption{keto: practitioner}
  \label{fig:prac_keto_des}
\end{subfigure}%
\begin{subfigure}{.5\columnwidth}
  \centering
  \includegraphics[width=\textwidth]{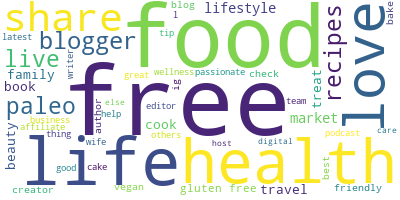}
  \caption{keto: promotional}
  \label{fig:promo_keto_des}
\end{subfigure}
\caption{Wordcloud for yoga and keto users' profile description.}
\label{fig:yoga_keto_desc_wc}
\end{figure}


\section{User Type Analysis}
In this section, we perform analysis on predicted user types from our data. To understand the topic of the user's tweets, we run LDA \cite{blei2003latent} based topic modeling over the tweets. Fig. \ref{fig:yoga_keto_twt_topic} shows the wordclouds of top $10$ keywords in each topic (we show $2$ topics here) from practitioner and promotional users' tweets both from yoga and keto dataset. For visualization purpose, we filter out the word `yoga' and `keto' because of the apparent high occurrences. Looking at yoga practitioners' topics, we notice that they tweet mostly about their yoga practice, yoga pose, yoga class, yoga practice time (Topic $0$ in Fig. \ref{fig:prac_yoga_twt_topic}). On the other hand, topic $1$ of practitioner mostly focus on practitioners' motivation of doing yoga, i.e., health and fitness benefit, spiritual connection, meditation. Topics from promotional yoga users mostly focus on joining yoga class, gym, fitness model, sport, yoga pant (Fig. \ref{fig:promo_yoga_twt_topic}). For keto practitioners, the tweets' topics are mostly related to ketogenic diet, low carb high fat food (lchf), ketonemia (presence of abnormally high concentration of ketone bodies in the blood), budget friendly ketodiet (Fig. \ref{fig:prac_keto_twt_topic}). Promotional keto users tweet mostly about keto recipe, paleo diet, gluten-free meal, supplements, anti-aging medicine (Fig. \ref{fig:promo_keto_twt_topic}).

We perform text analysis of user's profile description to understand what kind of words users use in their description which can distinguish their types based on their lifestyle. We create wordcloud with the most frequent words (Fig. \ref{fig:yoga_keto_desc_wc}) from yoga and keto users' profile description, filtering out the word `yoga' and `keto' respectively, because of the apparent high occurrences. We observe the words \textit\{`teacher', `instructor', `coach', `health', `meditation', `wellness', `writer', `yogi'\} in yoga practitioners' descriptions (Fig. \ref{fig:prac_yoga_des}). Promotional users have the following words in the description \textit\{`studio', `fitness', `health', `wellness', `class',  `offer', `community', `event', `business'\} (Fig. \ref{fig:promo_yoga_des}). Fig. \ref{fig:prac_keto_des} shows the wordcloud of the profile description of keto practitioner having words \textit\{`food', `nutrition', `coach', `author', `writer', `wife', `mom', `family'\}. Promotional users' description contains \textit\{`free', `food', `recipe', `health', `share', `blogger'\} words.
We notice that most users use \textit\{`love', `life', `live', `health'\} these words in their profile descriptions frequently, regardless of types.

\begin{figure}
\begin{subfigure}{1\columnwidth}
  \centering
  \includegraphics[width=\textwidth]{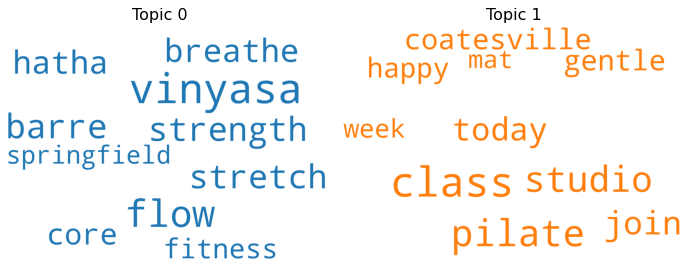}
  \caption{yoga: practitioner}
  \label{fig:prac_yoga_twt_topic_pos}
\end{subfigure}
\begin{subfigure}{1\columnwidth}
  \centering
  \includegraphics[width=\textwidth]{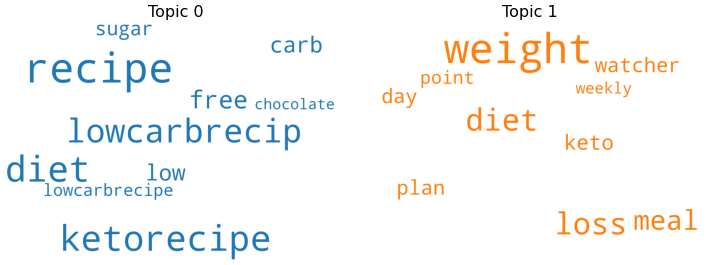}
  \caption{keto: practitioner}
  \label{fig:prac_keto_twt_topic_pos}
\end{subfigure}
\caption{Wordclouds of top $10$ keywords in each topic from yoga and keto practitioners' tweets having positive sentiment. Only $2$ topics from each dataset are shown here.}
\label{fig:prac_topic}
\end{figure}

\subsection{Users' Sentiment Analysis}
We analyze the sentiment of practitioners’ tweets
both for yoga and keto using VADER \cite{hutto2014vader} model which is used for text sentiment analysis that is sensitive to both polarity
(positive/negative) and intensity (strength) of emotion. 

For this analysis, we randomly select $350$ yoga practitioners and $150$ keto practitioners and choose their timeline tweets related to yoga and keto diet respectively. After running VADER sentiment analyzer, we obtain $9147$ positive sentiment and $853$ negative sentiment tweets for yoga practitioners. For keto practitioners, we have $7931$ positive and $2069$ negative sentiment tweets.

To understand the topic of the practitioners' tweets which have positive sentiment, we run LDA based topic modeling over the tweets. Fig. \ref{fig:prac_topic} shows the wordclouds of top $10$ keywords in each topic (we show $2$ topics here) from practitioners' tweets which have positive sentiment both from yoga (Fig. \ref{fig:prac_yoga_twt_topic_pos}) and keto (Fig. \ref{fig:prac_keto_twt_topic_pos}) dataset. 

For yoga practitioners positive sentiment tweets, Topic 0 (Fig. \ref{fig:prac_yoga_twt_topic_pos}) is  mostly related to `vinyasa' and `hatha' yoga. Vinyasa is a style of yoga characterized by stringing postures together so that we can move from one to another, seamlessly, using breath. It is commonly referred to as `flow' yoga. Hatha yoga allows for more stretching. Topic 1 is more about joining yoga class, studio. Looking at keto practitioners' (having positive sentiment) topics, we notice that they tweet mostly about the keto/lowcarb recipe, sugar free diet (Topic $0$ in Fig. \ref{fig:prac_keto_twt_topic_pos}). On the other hand, topic $1$ of practitioner mostly focus on practitioners' positive sentiment about weight loss, diet, meal plan. 


\section{Conclusion and Future Work}
In this paper, we propose a weakly supervised graph embedding based EM-style framework to characterize user types in social media. To demonstrate our model's effectiveness, we perform extensive experiments on real-world datasets focusing on `yoga' \& `keto diet' and our model outperforms the baselines. We perform data analysis on different user types from our data to show the topics from users' tweets and frequently used words in users' descriptions. Using our model with minimal tweaking in keywords selection for generating weak labels, we can distinguish any user types though they are not from the yoga/keto lifestyle communities. Our work can lead to new discussions on the analysis of health users' account types. While we focus specifically on lifestyle-related tweets, our approach is a general framework that can be adapted to other corpora. In the future, we aim to expand our work to detect communities based on different lifestyle decisions and understand their motivations.
\newline
Our code and the data are available here.\footnote[2]{\url{https://github.com/tunazislam/WeakSupervised-GraphEmb-UserRep}}

\section{Ethics Statement}
To the best of our knowledge no code of ethics
has been violated throughout the annotations and experiments done in this paper. We use illustrative examples to show the Twitter user profile descriptions, tweets in Fig. \ref{fig:keto_user}. As our dataset is comprised of tweets, we will share the original data based upon request for research purpose only.

\section*{Acknowledgments}
We gratefully acknowledge Md Masudur Rahman and the anonymous reviewers for their insightful comments. We further thank our annotators. This work was partially supported by an NSF CAREER award IIS-2048001.







\bibliography{ref}

\end{document}